\documentclass[final]{cvpr}

\usepackage{times}
\usepackage{epsfig}
\usepackage{graphicx}
\usepackage{amsmath}
\usepackage{amssymb}
\usepackage{multirow}
\usepackage{verbatim}
\usepackage{color} 
\usepackage{float}
\usepackage{bm}
\usepackage{enumitem}
\usepackage{booktabs}
\usepackage{subcaption}
\usepackage{makecell}
\usepackage{soul}
\usepackage{mathtools}
\usepackage{epigraph}
\usepackage{dsfont}
\usepackage[numbers,sort]{natbib} 
\DeclareMathOperator*{\argmax}{arg\,max}

\usepackage[pagebackref=true,breaklinks=true,letterpaper=true,colorlinks,bookmarks=false]{hyperref}
\hypersetup{
	plainpages=false,
	colorlinks=true,              %
	linkcolor=blue,               %
	anchorcolor=blue,             %
	citecolor=blue,               %
	filecolor=blue,               %
	urlcolor=blue,                %
	pdfview=FitH,                 %
	pdfstartview=FitH,            %
	pdfpagelayout=SinglePage      %
}

\usepackage[pagebackref=true,breaklinks=true,colorlinks,bookmarks=false]{hyperref}

\definecolor{mygray}{gray}{0.6}
\usepackage{pifont}
\newcommand{\cmark}{\ding{51}}%
\newcommand{\xmark}{\ding{55}}%

\def\cvprPaperID{393} 
\def\confYear{CVPR 2021}

\begin{document}

\title{Thinking Fast and Slow: Efficient Text-to-Visual Retrieval with Transformers}

\author{
	Antoine Miech\textsuperscript{1}\footnotemark[1]
	\quad\quad\quad
	Jean-Baptiste Alayrac\textsuperscript{1}\footnotemark[1]
	\\
		\quad\quad
	Ivan Laptev\textsuperscript{2}
	\quad\quad\quad\quad\quad
	Josef Sivic\textsuperscript{3}
	\quad\quad\quad
	Andrew Zisserman\textsuperscript{1,4}
	\\
	\small{$^1$\normalsize{DeepMind} \quad $^2$ENS/Inria \small \quad $^3$CIIRC CTU \quad $^4$VGG Oxford}
	\\
	\small{\texttt{\{miech,jalayrac\}@google.com}}
}

\maketitle

\begin{abstract}
Our objective is language-based search of large-scale image and video datasets. For this task, 
the approach that consists of independently mapping text and vision to a joint embedding space, a.k.a. dual encoders, is attractive as 
retrieval scales and is efficient for billions of images using  approximate nearest neighbour search.
An alternative approach of using vision-text transformers with cross-attention gives
considerable improvements in accuracy over the joint embeddings, but 
is often inapplicable in practice for large-scale retrieval given the cost of
the cross-attention mechanisms required for each sample at test time.
This work combines the best of both worlds. We make the following three contributions. 
First, we equip transformer-based models with a new fine-grained cross-attention architecture, providing  significant improvements in retrieval accuracy 
whilst preserving scalability.
Second, we introduce a generic approach for combining a {\em Fast} dual encoder model with our {\em Slow} but accurate transformer-based model via distillation and re-ranking.
Finally, we validate our approach on the Flickr30K {\em image} dataset where we show an increase in inference speed by several orders of magnitude while having results competitive to the state of the art.
We also extend our method to the video domain, improving the state of the art on the VATEX dataset.
\end{abstract}

\section{Introduction}

\begin{figure}[t]
\centering
\includegraphics[width=\linewidth]{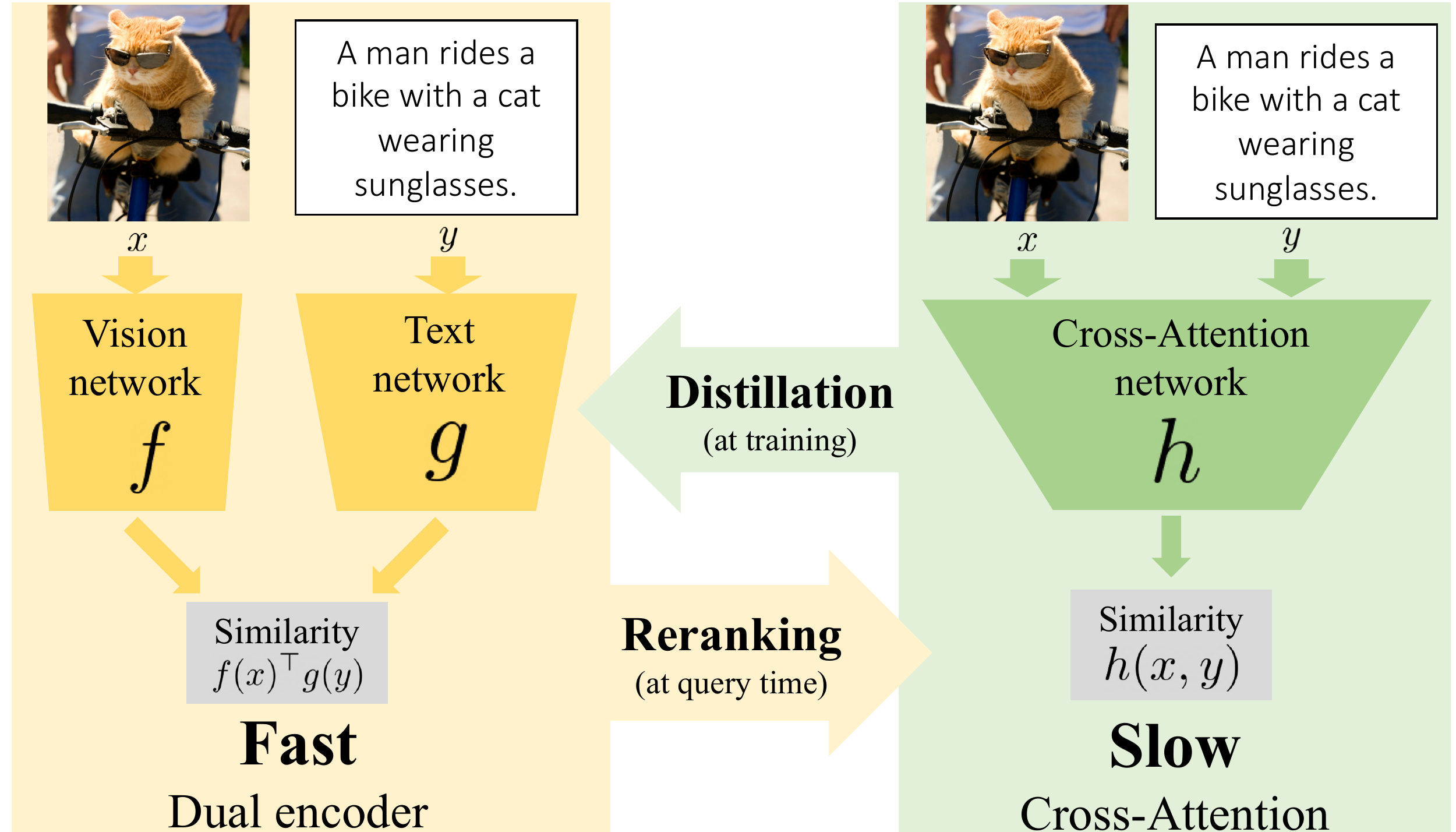}
\caption{\small{On the left, the \emph{Fast} models, a.k.a dual encoders, independently process the input image and text to compute a similarity score via a single dot product, which can be efficiently indexed and is thus amenable to large-scale search.
On the right, the \emph{Slow} models, a.k.a cross-attention models, jointly process the input image and text with cross-modal attention to compute a similarity score.
Fast and indexable models are improved by \emph{Slow} models via distillation at training time (offline).  \emph{Slow} models are accelerated and improved with the distilled \emph{Fast} approaches using a re-ranking strategy at query time. 
}}
\label{fig:fast_slow}
\end{figure}

\renewcommand{\thefootnote}{\fnsymbol{footnote}}
\footnotetext[1]{Equal contribution.}
\renewcommand*{\thefootnote}{\arabic{footnote}}
\footnotetext[1]{D\'{e}partement d'informatique de l'ENS, \'{E}cole normale sup\'{e}rieure, CNRS, PSL Research University, 75005 Paris, France.}
\footnotetext[3]{Czech Institute of Informatics, Robotics and Cybernetics at the Czech Technical University in Prague.}
\footnotetext[4]{VGG, Dept.\  of Engineering Science, University of Oxford}

Imagine yourself looking for an image that best matches a given textual description among thousands of other images. 
One effective way would be to first isolate a few promising candidates by giving a quick glance at all the images with a \emph{fast} process, \eg by eliminating images that have clearly nothing in common with the description.
In the second phase, you may start paying more attention to image details with a \emph{slow} process, \eg by grounding individual words of a query sentence to make sure the scrutinized image is the best match.

Analogous to the \emph{fast} process above, fast retrieval systems can be implemented by separately encoding visual and textual inputs into a joint embedding vector space where similarities can be computed by dot product.
Such methods are regarded as \emph{indexable}, \ie they allow application of fast approximate nearest neighbour search~\cite{datar2004locality,muja2009fast,sivic03videogoogle,jegou2010product} and enable efficient billion-scale image retrieval.
However, the accuracy of such methods is limited due to the simplicity of vision-text interaction model defined by the dot product in the joint embedding space.
We refer to these techniques as \emph{Dual Encoders (DE) or Fast approaches}.

Vision-text transformers compare each word to all locations in the image using cross-attention~\cite{lu2019vilbert,desai2020virtex,huang2020pixel}, 
allowing for grounding, and can be related to the \emph{slow} process mentioned earlier.
Such methods, referred to here as \emph{Cross-attention (CA) or Slow approaches}, significantly boost retrieval performance.
Modeling text-vision interactions with attention, however, makes these models slow and impractical for large-scale image retrieval given the cost of
the cross-attention mechanisms required for each sample at test time.
Hence, the challenge we consider is the following: How to benefit from accurate cross-attention mechanisms while preserving the fast and scalable visual search?

Our short answer is: By \emph{thinking Fast and Slow~\cite{daniel2017thinking}.}
As illustrated in Figure~\ref{fig:fast_slow},  we propose to combine dual encoder approaches with cross-attention via two complementary mechanisms.
First, we improve \emph{Fast} DE models with a novel \emph{distillation} objective that transfers knowledge from accurate but \emph{Slow} CA models to the \emph{Fast} and indexable dual encoders.
Second, we propose to combine DE and CA models with re-ranking where a few most promising candidates obtained with the \emph{Fast} model are re-ranked using the \emph{Slow} model.
Our resulting approach is both fast and accurate.
Since the speed of CA is not a bottleneck anymore, we further improve performance by enriching the vision-text cross-attention model with a novel feature map upsampling mechanism enabling fine-grained attention.
Note that our work can also be applied to vision-to-text retrieval.
However, we focus on text-to-vision retrieval due to its wider practical application.

\noindent \textbf{Contributions.}
\textbf{(i)}~We first propose a gradual feature upsampling architecture for improved and fine-grained vision and text cross-attention.
Our model is trained with a bi-directional captioning loss which is remarkably competitive for retrieval compared to standard cross-modal matching objectives.
\textbf{(ii)}~We introduce a generic approach for scaling-up transformer-based vision-text retrieval using two core ideas:
a method to distill the knowledge of \emph{Slow} cross-attention models into \emph{Fast} dual-encoders, and 
re-ranking top results of the {\em Fast} models with the \emph{Slow} ones.
\textbf{(iii)}~Finally, we validate our approach on image retrieval with the COCO~\cite{lin14coco} and Flickr30K~\cite{plummer2015flickr30k} datasets and show we can reduce the inference time of powerful transformer-based models by 100$\times$ whilst also getting competitive results to the state of the art.
We also successfully extend our approach to text-to-video retrieval and improve state of the art on the challenging VATEX~\cite{wang2019vatex} dataset.

\section{Related work}

\noindent \textbf{Vision and Language models.} 
Driven by the significant advances in language understanding lead by Transformers~\cite{devlin2019bert,vaswani2017attention}, recent works have explored the use of these architectures for vision and language tasks.
Many of them in image~\cite{chen2019uniter,lu2019vilbert,li2019unicodervl,li2019visualbert,li2020oscar,tan2019lxmert,su2019vl,zhou2020unified} or video~\cite{zhu2020actbert} rely on pretrained object detectors used for extracting ROIs that are viewed as individual visual words.
A few other works, such as PixelBERT~\cite{huang2020pixel} and VirTex~\cite{desai2020virtex} for images or HERO~\cite{li2020hero} for video,
operate directly over dense feature maps instead of relying on object detectors.
In these approaches, both vision and text inputs are fed into a Transformer-based model usually pretrained with multiple losses such as a cross-modal matching loss, a masked language modelling or a masked region modelling loss.
Other non-Transformer based vision and text approaches used recurrent neural networks~\cite{donahue14long,dong19dual,klein15associating}, MLP~\cite{wang2018learning,wang2016learning}, or bag-of-words~\cite{gong14multi,miech19howto100m} text models. 
These models are then usually optimized with objectives such as CCA~\cite{gong14multi}, max margin triplet loss~\cite{dong19dual,wang2018learning,wang2016learning,wray2019fine,wu2017sampling}, contrastive loss~\cite{gupta2020grounding} and, more related to our work, by maximizing text log-likelihoods conditioned on the image~\cite{donahue14long}.
In our work, we focus on the powerful vision-text Transformer models for retrieval and particularly address their scalability, which was frequently neglected by prior work. 

\noindent \textbf{Language-based visual search.}
A large number of vision and language retrieval models~\cite{dong19dual,gong14multi,gong14improving,klein15associating,miech18learning,miech19howto100m,pan16jointly,plummer2017enhancing,xu2015jointly,
wang2018learning,wang2016learning,wray2019fine,wu2017sampling} use a dual encoder architecture where the text and vision inputs are separately embedded into a joint space.
These approaches can efficiently benefit from numerous approximate nearest neighbour search methods such as: product quantization~\cite{jegou2010product}, inverted indexes~\cite{sivic03videogoogle}, hierarchical clustering~\cite{muja2009fast} or locality sensitive hashing~\cite{datar2004locality}, for fast and scalable visual search.
In contrast, state-of-the-art retrieval models rely on large vision-text multimodal transformers~\cite{chen2019uniter,huang2020pixel,li2019unicodervl,li2019visualbert,lu2019vilbert,lu202012,su2019vl,tan2019lxmert,zhou2020unified}.
In these approaches, both vision and text inputs are fed into a cross-modal attention branch to compute the similarity between the two inputs.
This scoring mechanism based on cross-modal attention makes it particularly inadequate for indexing and thus challenging to deploy at a large scale.
Our work aims at addressing this issue by connecting scalable visual search techniques with these powerful yet non-indexable vision-text cross-attention based models.

\noindent \textbf{Re-ranking.}
Re-ranking retrieval results is standard in retrieval systems.
In computer vision, the idea of geometric verification~\cite{Philbin07,Jegou08} is used in object retrieval to re-rank objects that better match the query given spatial consistency criteria.
Query expansion~\cite{Chum07b} is another re-ranking technique where the query is reformulated given top retrieved candidates, and recent work has brought attention mechanisms into deep learning methods for query expansion~\cite{Gordo20}.
Related to language-based visual search, re-ranking by a video-language temporal alignment model has been used to improve efficient moment retrieval in video~\cite{escorcia2019temporal}. 
In contrast, we focus on transformer-based cross-attention models and develop a distillation objective for
efficient retrieval. 

\noindent \textbf{Distillation.}
Knowledge distillation~\cite{hinton2015distilling,bucilua2006model} has proven to be effective for improving performance in various computer vision domains such as weakly-supervised learning~\cite{li2017learning,Radosavovic_2018_CVPR}, depth estimation~\cite{gupta2016cross}, action recognition~\cite{stroud2020d3d}, semantic segmentation~\cite{Liu_2019_CVPR}, self-supervised learning~\cite{piergiovanni2020evolving} or self-training~\cite{xie2020self}.
One major application of distillation is in compressing large and computationally expensive models in 
language analysis~\cite{sanh2019distilbert}, object detection~\cite{chen2017learning},  image classification or speech recognition~\cite{hinton2015distilling} into smaller and computationally less demanding models. 
In this work, we describe a distillation mechanism for the compression of powerful but non-indexable vision-text models into indexable models suitable for efficient retrieval. 

\section{Thinking Fast and Slow for Retrieval}
\label{sec:approach}

This section describes our proposed approach to learn both fast and accurate model for language-based image retrieval.

Our goal is to train the model to output a similarity score between an input image $x$ and a textual description $y$.
In this work, we focus on two families of models: the \emph{Fast} and the \emph{Slow} models, as illustrated in Figure~\ref{fig:fast_slow}.

The \emph{Fast} model, referred to as the \emph{dual encoder approach}, consists of extracting modality-specific embeddings: $f(x)\in\mathbb{R}^d$  for the image and $g(y)\in\mathbb{R}^d$ for the text.
The core property of this approach is that the similarity between an image $x$ and a text $y$ can be computed via a single dot product $f(x)^\top g(y)$.
Hence, these methods can benefit from approximate nearest neighbour search for efficient large-scale retrieval~\cite{jegou2010product,sivic03videogoogle,muja2009fast}.

The \emph{Slow} model, referred to as the \emph{cross-attention} approach differs by a more complex modality merging strategy based on cross-modal attention.
We assume the given similarity score $h(x, y)$ cannot be decomposed as a dot product and as such is not indexable.
These models allow for richer interactions between the visual and textual representations, which leads to better scoring mechanisms, though at a higher computational cost.

Section~\ref{sec:im_text_model} introduces the (\emph{Slow}) cross-attention model considered in this work and details our contribution on the model architecture that leads to a more accurate text-to-image retrieval system.
Section~\ref{sec:fast_and_slow} describes how we obtain both a \emph{fast} and \emph{accurate} retrieval method by combining the advantages of the two families of models.

\subsection{Thinking Slow with cross-attention}
\label{sec:im_text_model}
Given an image $x$ and a text description $y$, a \emph{Slow} cross-attention retrieval model $h$ computes a similarity score between the image and text as:
\begin{equation}
\label{eq:cross_attention_models}
h(x,y) = A(\phi(x), y),
\end{equation}
where $\phi$ is a visual encoder (\eg a CNN).
$A$ is a network that computes a similarity score between $\phi(x)$ and $y$ using cross-attention~\cite{lu2019vilbert,vaswani2017attention} mechanisms, \ie the text attends to the image or vice versa via multiple non-linear functions involving both the visual and language representations.
Such models emulate a \emph{slow} process of attention which results in better text-to-image retrieval.

We propose two important innovations to improve such models.
First, we introduce a novel architecture that enables fine-grained visual-text cross-attention by efficiently increasing the resolution of the attended high-level image features.
Second, we propose to revisit the use of a captioning loss~\cite{donahue14long} to train retrieval models and discuss the benefits over standard alternatives that use classification or ranking loss~\cite{chen2019uniter,lu2019vilbert,li2019unicodervl,li2019visualbert,tan2019lxmert,su2019vl,zhou2020unified}.

\noindent
\begin{figure*}[t]
\centering
\includegraphics[width=\linewidth]{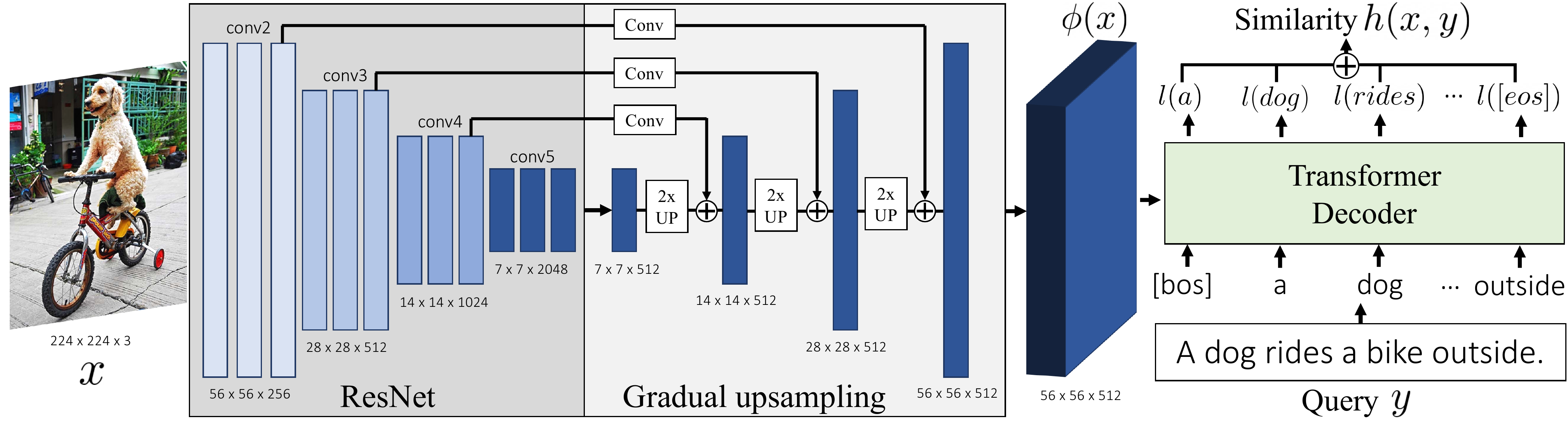}
\caption{\small{{\bf Our Slow retrieval model} computes a similarity score $h(x,y)$ between image $x$ and query text description $y$ by estimating the log-likelihood of $y$ conditioned on $x$. 
In other words, given an input text query $y$, we perform retrieval by searching for an image $x$ that is the most likely to decode caption $y$.
$l(.)$ denotes the log probability of a word given preceding words and the image.
The decoder is a Transformer that takes as the conditioning signal a high-resolution (here $56 \times 56$) feature map $\phi(x)$. 
In this example, $\phi(x)$ is obtained by gradually upsampling the last convolutional layer of ResNet ($7 \times 7$) while incorporating features from earlier high-resolution feature maps.
The decoder performs bidirectional captioning but, 
for the sake of simplicity, we only illustrate here the forward decoding transformer.}}
\label{fig:architecture}
\end{figure*}

\textbf{A novel architecture for fine-grained vision-text cross-attention.}
A typical approach to attend to visual features produced by a CNN is to consider the last convolutional layer~\cite{desai2020virtex,huang2020pixel}. 
The feature map is flattened into a set of feature vectors that are used as input to vision-language cross-attention modules.
For example, a $224\times 224$ input image passed through a ResNet-50~\cite{he16resnet} outputs a $7\times 7$ feature map that is flattened into $49$ vectors.
While the last feature map produces high-level semantic information crucial for grounding text description into images, this last feature map is also severely downsampled.
As a result, useful fine-grained visual information for grounding text descriptions might be lost in this downsampling process.

One solution to the problem is to increase the input image resolution.
However, this significantly raises the cost of running the visual backbone.
Inspired by previous work in segmentation~\cite{badrinarayanan2017segnet,hariharan2015hypercolumns,ronneberger2015u} and human pose estimation~\cite{newell2016stacked}, we instead propose to gradually upsample the last convolutional feature map conditioned on earlier higher resolution feature maps, as illustrated in~Figure~\ref{fig:architecture}.
We choose a lightweight architecture for this upsampling process inspired by recent advances in efficient object detection~\cite{tan2020efficientdet}.
In Section~\ref{sec:experiments}, we show large improvements of this approach over several baselines and also show its complementarity to having higher resolution input images, clearly demonstrating the benefits of the proposed fine-grained vision-language cross-attention.

\noindent \textbf{Bi-directional captioning objective for retrieval.}
A majority of text-vision transformer-based retrieval models~\cite{chen2019uniter,lu2019vilbert,li2019unicodervl,li2019visualbert,tan2019lxmert,su2019vl,zhou2020unified} rely on a cross-modal image-text matching loss to discriminate positive image-text pairs $(x,y)$ from negative ones. 
In this work, we instead explore the use of a captioning model for retrieval.
Given an input text query $y$, retrieval can be done by searching the image collection for the image $x$ that leads to the highest likelihood of $y$ given $x$ according to the model.
In detail, we take inspiration from VirTex~\cite{desai2020virtex} and design the cross-attention module $A$ as a stack of Transformer decoders~\cite{vaswani2017attention} taking the visual feature map $\phi(x)$ as an encoding state.
Each layer of the decoder is composed of a masked text self-attention layer, followed by a cross-attention layer that enables the text to attend to the visual features and finally a feed forward layer.
One advantage of this architecture compared to standard multimodal transformers~\cite{chen2019uniter,lu2019vilbert,li2019unicodervl,li2019visualbert,tan2019lxmert,su2019vl,zhou2020unified} is the absence of self-attention layers on visual features, which allows the resolution of the visual feature map $\phi(x)$ 
to be scaled to thousands of vectors. 
We write the input text as $y=[y^1,\dots,y^L]$ where $L$ is the number of words.
Formally, the model $h$ scores a pair of image and text $(x,y)$ as:
\begin{equation}
\label{eq:ca_output}
h(x, y) = h_{fwd}(x, y) + h_{bwd}(x, y),
\end{equation}
where $h_{fwd}(x, y)$ (resp. $h_{bwd}(x, y)$) is the forward (resp. backward) log-likelihood of the caption $y$ given the image $x$ according to the model:
\begin{equation}
\label{eq:forward}
h_{fwd}(x, y) = \sum_{l=1}^{L} \log(p(y^l|y^{l-1},\dots,y^1,\phi(x); \theta_{fwd})),
\end{equation}
where $p(y^l|y^{l-1},\dots,y^1,\phi(x);\theta)$ corresponds to the output probability of a decoder model parametrized by $\theta$ for the token $y^l$ at position $l$ given the previously fed tokens $y^{l-1},\dots,y^1$ and the encoded image $\phi(x)$.
$\theta_{fwd}$ is  the parameters of the forward transformer models. 
$h_{bwd}(x, y)$ is the same but with the sequence $y^1,\dots,y^{L}$  in reverse order.

The parameters of the visual backbone, the forward and backward transformer models are obtained by minimizing $\mathcal{L}_{\text{CA}}  = - \sum_{i=1}^{n} h(x_i, y_i)$ where $n$ is the number of annotated pairs of images and text descriptions $\{(x_i, y_i)\}_{i \in [1, n]}$.

We show in Section~\ref{sec:experiments} that models trained for captioning can perform on-par with models trained with the usual contrastive  image-text matching loss.
At first sight this may appear surprising as the image-text matching loss seems more suited for retrieval,  notably because it explicitly integrates negative examples.
However, when looked at more closely, the captioning loss actually shares similarities with a contrastive loss: for each ground truth token of the sequence a cross entropy loss is taken (see Eq.~\eqref{eq:forward}) which effectively means that all other tokens in the vocabulary are considered as negatives.

In this section, we have described the architecture and the chosen loss for training our accurate \emph{Slow} cross-attention model for retrieval.
One key remaining challenge is in the scaling of $h(x,y)$ using Eq.~\eqref{eq:cross_attention_models} to large image datasets as: \textit{(i)}~the network $A$ is expensive to run and \textit{(ii)}~the resulting intermediate encoded image, $\phi(x)$, is too large to fit the entire encoded dataset in memory.
Next, we introduce a generic method, effective beyond the scope of our proposed \emph{Slow} model, for efficiently running such cross-modal attention-based models at a large scale.

\subsection{Thinking Faster and better for retrieval}
\label{sec:fast_and_slow}

In this section, we introduce an approach to scale-up the \emph{Slow} transformer-based cross-attention model, described in the previous section, using two complementary ideas. First, we distill the knowledge of the \emph{Slow} cross-attention model into a \emph{Fast} dual-encoder model that can be efficiently indexed. Second, we combine the \emph{Fast} dual-encoder model with the \emph{Slow} cross-attention model via a re-ranking mechanism. The outcome is more than $100\times$ speed-up and, interestingly, an improved retrieval accuracy of the combined \emph{Fast and Slow} model.  Next, we give details of the \emph{Fast} dual encoder model, then explain the distillation of the \emph{Slow} model into the \emph{Fast} model using a teacher-student approach, and finally describe the re-ranking mechanism to combine the outputs of the two models.
Because our approach is model agnostic, the \emph{Slow} model can refer to any vision-text transformer and the \emph{Fast} to any dual-encoder model.
An overview of the approach is illustrated in Figure~\ref{fig:fast_slow}.

\paragraph{Fast indexable dual encoder models.}
We consider \emph{Fast}  dual encoder models, that extract modality specific embeddings: $f(x)\in\mathbb{R}^d$  from image $x$,  and $g(y)\in\mathbb{R}^d$ from text $y$. The core property of this approach is that the similarity between the embedded image $x$ and text $y$ is measured with a dot product, $f(x)^\top g(y)$. The objective is to learn embeddings $f(x)$ and $g(y)$ so that semantically related images and text have high similarity and the similarity of unrelated images and text is low. To achieve that we train these embeddings by minimizing the standard noise contrastive estimation (\textbf{NCE})~\cite{gutmann2010noise,jozefowicz2016exploring} objective:
\begin{equation}
\label{eq:contrastive_objective}
\mathcal{L}_{\text{DE}}  = -\sum_{i=1}^n\log\left(\frac{e^{f(x_i)^\top g(y_i)}}{e^{f(x_i)^\top g(y_i)}+\sum\limits_{(x',y')\in\mathcal{N}_i}e^{f(x')^\top g(y')}}\right),
\end{equation}
which contrasts the score of the positive pair $(x_i, y_i)$ to a set of negative pairs sampled from a negative set $\mathcal{N}_i$.
In our case, the image encoder $f$ is a globally pooled output of a CNN while the text encoder $g$ is either a bag-of-words~\cite{miech19howto100m} representation or a more sophisticated BERT~\cite{devlin2019bert} encoder.
Implementation details are provided in Section~\ref{sec:datasets}.

\paragraph{Distilling the \emph{Slow} model into the \emph{Fast} model.}

Given the superiority of cross-attention models over dual encoders for retrieval, we investigate how to distill~\cite{hinton2015distilling} the knowledge of the cross-attention model to a dual encoder.  
To achieve that we introduce a novel loss. 

In detail, the key challenge is that, as opposed to standard distillation used for classification models, here we do not have a small finite set of classes but potentially an infinite set of possible sequences of words describing an image. Therefore, we cannot directly apply the standard formulation of distillation proposed in~\cite{hinton2015distilling}.

To address this issue, we introduce the following extension of distillation for our image-text setup. 
Given an image-text pair $(x_i, y_i)$, we sample a finite subset of image-text pairs $\mathcal{B}_i=\{(x_i, y_i)\} \cup \{(x, y_i) |  \ x \neq x_i \} $, where we construct additional image-text pairs with the same text query $y_i$ but different images $x$.
Note that this is similar to the setup that would be used to perform retrieval of images $x$ given a text query $y_i$.
In practice, we sample different images $x$ within the same training batch.
We can write a probability distribution measuring the likelihood of the pair $(x,y) \in \mathcal{B}_i $ according to the \emph{Slow} teacher model $h(x,y)$ (given by eq. \eqref{eq:cross_attention_models}) over subset $\mathcal{B}_i$ as:
\begin{equation}
\label{eq:teacher_target}
p(\mathcal{B}_i)(x,y) = \frac{\exp(h(x, y)/\tau)}{\sum_{(x', y')\in{\mathcal{B}_i}}\exp(h(x', y')/\tau)},
\end{equation}
where $\tau > 0$ is a temperature parameter controlling the smoothness of the distribution.
We can obtain a similar distribution from the \emph{Fast} student model, by replacing $h(x,y)$ from Eq.~\eqref{eq:teacher_target} by $f(x)^\top g(y)$:
\begin{equation}
q(\mathcal{B}_i)(x,y)= \frac{\exp(f(x)^\top g(y)/\tau)}{\sum_{(x', y')\in\mathcal{B}_i}\exp(f(x')^\top g(y')/\tau)}.
\end{equation}

Given the above definition of the sampled distributions, we use the following distillation loss that measures the similarity between the teacher distribution $p(\mathcal{B}_i)$ and the student distribution $q(\mathcal{B}_i)$ as :
\begin{equation}
\label{eq:distill_objective}
\mathcal{L}_{\text{distill}} = \sum_{i=1}^n \mathcal{H}(p(\mathcal{B}_i), q(\mathcal{B}_i	)),
\end{equation}
where $\mathcal{H}$ is the cross entropy between the two distributions.
The intuition is that the teacher model provides soft targets over the sampled image-text pairs as opposed to binary targets in the case of a single positive pair and the rest of the pairs being negative. 
Similarly to the standard distillation~\cite{hinton2015distilling}, we combine the distillation loss~\eqref{eq:distill_objective} with the DE loss~\eqref{eq:contrastive_objective} weighted with $\alpha > 0$ to get our final objective as:

\begin{equation}
\label{eq:final_objective}
\min_{f,g}  \mathcal{L}_{\text{distill}} + \alpha \mathcal{L}_{\text{DE}}.
\end{equation}

\paragraph{Re-ranking the \emph{Fast} results with the \emph{Slow} model.}
The distillation alone is usually not sufficient to recover the full accuracy of the \emph{Slow}  model using the \emph{Fast} model.
To address this issue, we use the \emph{Slow}  model at inference time to re-rank a few of the top retrieved candidates obtained using the \emph{Fast} model.
First, the entire dataset is ranked by the (Distilled) \emph{Fast} model that can be done efficiently using approximate nearest neighbour search, which often has only sub-linear complexity in the dataset size.
Then the top $K$ (\eg 10 or 50) results are re-ranked by the \emph{Slow} model.
As the \emph{Slow} model is applied only to the top $K$ results its application does not depend on the size of the database.

More precisely, given an input text query $y$ and an image database $\mathcal{X}$ containing a large number of $m$ images, we first obtain a subset of $K$ images $\mathcal{X}_K$ (where $K \ll m$) that have the highest score according to the \emph{Fast} dual encoder model.
We then retrieve the final top ranked image by re-ranking the candidates using the \emph{Slow} model:
\begin{equation}
\label{eq:rerank}
\argmax_{x\in{\mathcal{X}_K}} h(x, y) + \beta f(x)^\top g(y) ,
\end{equation}
where $\beta$ is a positive hyper-parameter that weights the output scores of the two models.
In the experimental Section~\ref{sec:experiments}, we show that \emph{combined with distillation}, re-ranking less than ten examples out of thousands can be sufficient to recover the performance of the \emph{Slow} model.

\section{Experiments}
\label{sec:experiments}

In this section, we evaluate the benefits of our approach on the task of text-to-vision retrieval.
We describe the datasets and baselines used for evaluation in Section~\ref{sec:datasets}.
In Section~\ref{sec:ca_vs_de} we validate the advantages of cross-attention models with captioning objectives as well as our use of gradually upsampled features for retrieval.
Section~\ref{sec:rerank} evaluates the benefit of the distillation and re-ranking.
In Section~\ref{sec:sota}, we compare our approach to other published state-of-the-art retrieval methods in the image domain and
show state of the art results in the video domain.

\subsection{Datasets and models}
\label{sec:datasets}

\noindent \textbf{MS-COCO~\cite{lin14coco}.} We use this image-caption dataset for training and validating our approach.
We use the splits of~\cite{chen2015microsoft} (118K/5K images for train/validation with 5 captions per image).
We only use the first caption of each image to make validation faster for slow models.
C-R@1 (resp. C-R@5) refers to recall at 1 (resp. 5) on the validation set.

\noindent \textbf{Conceptual Captions (CC)~\cite{sharma2018conceptual}.} 
We use this dataset for training our models (2.7M training images (out of the 3.2M) at the time of submission).
CC contains images and captions automatically scraped from the web which shows our method can work in a weakly-supervised training regime.

\noindent \textbf{Flickr30K~\cite{plummer2015flickr30k}.} 
We use this dataset for zero-shot evaluation (\ie we train on  COCO or CC and test on Flickr) in the ablation study, as well as fine-tuning when comparing to the state of the art.
We use the splits of~\cite{karpathy14deepvisual} (29K/1014/1K for train/validation/test with 5 captions per image).
We report results on the validation set except in Section~\ref{sec:sota} where we report on the test split.
We abbreviate F-R@1 (resp.\ F-R@5) as the R@1 (resp.\ R@5) scores on Flickr.

\noindent \textbf{VATEX~\cite{wang2019vatex}.} 
VATEX contains around 40K short $10$ seconds clip from the Kinetics-600
dataset~\cite{carreira2018short} annotated with multiple descriptions.
In this work, we only use the $10$ English captions per video clip and ignore the additional Chinese captions.
We use the retrieval setup and splits from~\cite{chen2020fine}.

\noindent \textbf{Models.} 
For each model, the visual backbone is a ResNet-50 v2 CNN~\cite{he2016identity} trained from scratch.
Inputs are 224 $\times$ 224 crops for most of the validation experiments  unless specified otherwise.
Models are optimized with ADAM~\cite{kingma15adam},  and a cosine learning rate decay~\cite{loshchilov2016sgdr} with linear warm-up is employed for the learning rate.
The four main models used in this work are described next.

\noindent \textbf{NCE BoW} is a dual-encoder (DE) approach where the text encoder is a bag-of-words~\cite{miech19howto100m} on top of word2vec~\cite{mikolov13distributed} pretrained embeddings. 
The model is trained with the NCE\; loss given in Eq.~\eqref{eq:contrastive_objective} where the negative set $\mathcal{N}_i$ is constructed as in~\cite{miech2019end2end}.
We refer to \textbf{NCE BoW} as the \emph{Fast} approach.

\noindent \textbf{NCE BERT} is a DE approach where the text encoder is a pretrained BERT base model~\cite{devlin2019bert}. 
We take the [CLS] output for aggregating the text representation.
The model is also trained with the NCE loss given in Eq.~\eqref{eq:contrastive_objective}.

\noindent  \textbf{VirTex~\cite{desai2020virtex}} is a cross-attention (CA) based approach that originally aims at learning visual representations from text data using a
captioning pretext task.
We chose this visual captioning model as another point of comparison for the effectiveness of Transformer-based captioning models for text-to-vision retrieval.

\noindent  \textbf{PixelBERT~\cite{huang2020pixel}} is a CA approach trained with the standard masked language modelling (MLM) and image-text matching (ITM) losses for retrieval.
One difference between our implementation and the original PixelBERT is the use of 224 $\times$ 224 images for a fair comparison with other models.
Note that the main difference with VirTex is in the vision-text Transformer architecture: PixelBERT uses a deep 12-layer Transformer encoder while VirTex uses a shallow 3-layer Transformer decoder to merge vision and language.

We chose PixelBERT and VirTex for their complementarity and their simplicity since they do not rely on object detectors.
We reimplemented both methods so that we could ensure that they were comparable.
Next, we describe the details of our proposed CA approach.

\noindent  \textbf{\emph{Slow} model architecture.}
For the upsampling, we follow a similar strategy as used in BiFPN~\cite{tan2020efficientdet}.
For the decoder, we use a stack of 3 Transformer decoders with hidden dimension 512 and 8 attention heads.
Full details about the architecture are provided in Appendix~\ref{app:arch}.

\begin{table}
\resizebox{\linewidth}{!}{	
      \begin{tabular}{@{}lcccccc@{}}
	Model & Type &  Train  & F-R@1 & F-R@5  & C-R@1 & C-R@5  \\
	\midrule
	\emph{Fast} NCE BoW & \multirow{2}{*}{DE} &  \multirow{2}{*}{COCO}   &  27.2 & 54.1  & 24.8  & 53.7 \\
	NCE BERT & &                                                                           &  24.4 & 48.0 & 24.2 & 52.0 \\
	\midrule
	PixelBERT & \multirow{3}{*}{CA}  &   \multirow{3}{*}{COCO}  &  30.0     & 55.1     & 25.1       &  52.5 \\
	VirTex Fwd only& &                                                                                  &   33.4     &  58.1 & 31.8  & 61.2 \\
	VirTex & &                                                                                  &   \textbf{38.1}     &  \textbf{62.8}  & \textbf{35.1}  & \textbf{64.6} \\
	\midrule
	\emph{Fast} NCE BoW & \multirow{2}{*}{DE} &\multirow{2}{*}{CC}         &  32.4  & 59.6 & 14.9   & 35.0  \\
	NCE BERT & &                                                                         &   25.8 &  50.7  & 12.2  & 29.8 \\
	\midrule
	PixelBERT & \multirow{3}{*}{CA} &  \multirow{3}{*}{CC}  &   30.4    &  57.7   & 14.1   & 33.6    \\
	VirTex Fwd only& &                                                                                &   32.2 & 58.4  &  14.7    & 32.9 \\
	VirTex & &                                                                                &   \textbf{35.0} & \textbf{60.7}   &  \textbf{16.1}    & \textbf{36.4} \\
\end{tabular}	
}
\caption{ \small Dual encoder (DE) and Cross-attention (CA) comparison. 
F-R@K corresponds to the recall at K on Flickr while C-R@K is the recall at K on COCO.}\label{tab:ablation_model}
\end{table}

\begin{table}
	\resizebox{\linewidth}{!}{	
		\begin{tabular}{@{}lcccccc@{}}
			Feature map & Size & F-R@1 & F-R@5  & C-R@1 & C-R@5  \\
			\midrule
			\emph{Slow} 96x96  &  384 &  \textbf{44.8} & \textbf{70.5} & \textbf{39.0} & \textbf{67.7} \\
			\emph{Slow} 56x56  &   \multirow{3}{*}{224}  &  42.2 & 66.8 & 38.5 & 65.2 \\
			\emph{Slow} 28x28   &  &  40.4 &  66.3  & 37.4  & 66.8  \\
			\emph{Slow} 14x14   &   & 39.2 & 63.8  & 36.8  & 64.9  \\
			\hline
			VirTex \texttt{conv5} (7x7) &  \multirow{4 }{*}{224} &  38.1     &  62.8  & 35.1  & 64.6 \\
			VirTex \texttt{conv4} (14x14) &   & 38.9  &  64.4  & 34.9 & 63.5 \\
			VirTex \texttt{conv3} (28x28) &   & 32.4     &  57.9  & 30.4 & 58.3 \\
			VirTex \texttt{conv2} (56x56) &  &  20.6    &  41.1  & 18.3  & 43.0 \\
		\end{tabular}	
	}
	\caption{ \small Gradual upsampling with different feature map size.  Size denotes the input image size.  Models are trained on COCO.}\label{tab:ablation_upsample}
\end{table}

\subsection{Improving cross-attention for retrieval}
\label{sec:ca_vs_de}

In this section, we provide an experimental study on the use of cross-attention models for retrieval.
All our results are validated on the COCO and the Flickr30K validation sets with models pretrained on COCO and CC training sets.
Our main findings are summarized below.

\noindent 
\textbf{Cross-attention models are better than Dual Encoders.}
Table~\ref{tab:ablation_model} compares various approaches for retrieval.
We observe that cross-attention models (PixelBERT and the VirTex variants), overall, outperform the dual encoders (NCE BoW and BERT).
Interestingly, using a simple BoW text encoder performs better than using a BERT text encoder for the DE models.
This suggests that the complexity of the language model is not the key factor for good performance but instead that complex merging strategy obtained from text-vision cross-attention may matter most for retrieval.

\noindent 
\textbf{Captioning models are surprisingly good for retrieval.}
Comparing `PixelBERT' against the `VirTex Fwd only' in Table~\ref{tab:ablation_model} with the exact same input dimensions and visual backbones, we see that using a captioning loss leads to better results than using an image-text matching loss coupled with a masked language modelling loss.
Backward captioning further improves retrieval performance.
This result demonstrates that captioning can be a strong alternative to the usual image-text matching losses for retrieval.

\noindent 
\textbf{Benefits of our gradual upsampling architecture design.}
In Table~\ref{tab:ablation_upsample}, we provide the results using the proposed upsampling strategy for our \emph{Slow} model presented in Section~\ref{sec:im_text_model} and illustrated in Figure~\ref{fig:architecture}.
We observe significant improvements over the VirTex baseline, denoted with \texttt{conv5} (7x7), (more than 4\% for R@1 on Flickr and more than 3\% on COCO) for the largest upsampling $56\times56$.
We also confirm that the performance gap does not just come from having a larger input feature map to attend to as the baseline with the output of ResNet \texttt{conv2}, which has a resolution of $56\times56$, performs poorly.
We believe it is important to keep high-level abstraction in the feature maps while having high resolution which our proposed architecture allows. 
It is also important to highlight that the proposed architecture leads to our best performing model and can be combined with higher input resolution for further improvements.
However, our proposed changes increase the inference time.
Next, we explore how to recover the speed.

\begin{table}
\resizebox{\linewidth}{!}{	
      \begin{tabular}{@{}llccccc@{}}
	Student & Teacher &  Train  & F-R@1 & F-R@5  & C-R@1 & C-R@5  \\
	\midrule
	\multirow{2}{*}{\emph{Fast}}  & None &  \multirow{3}{*}{COCO}   &  27.2 & 54.1  & 24.8  & 53.7  \\
	 & \emph{Slow} &                                          &   \textbf{37.7}    &  \textbf{64.7}  &  \textbf{32.5}      & \textbf{62.1} \\
	\multicolumn{2}{c}{\textcolor{mygray}{\emph{Slow} upper bound}}   &                           &  \textcolor{mygray}{42.2} & \textcolor{mygray}{66.8} & \textcolor{mygray}{38.5} & \textcolor{mygray}{65.2} \\
	\midrule
	\multirow{2}{*}{\emph{Fast}} & None &  \multirow{3}{*}{CC}   &  32.4  & 59.6 & 14.9   & 35.0    \\
	  & \emph{Slow}  &                                          &  \textbf{33.4}  &  \textbf{60.1}  & \textbf{17.2}   & \textbf{38.1}  \\
	\multicolumn{2}{c}{\textcolor{mygray}{\emph{Slow} upper bound}}   &                       &  \textcolor{mygray}{41.7} &  \textcolor{mygray}{67.5}  &  \textcolor{mygray}{19.8}   & \textcolor{mygray}{40.9}  \\
\end{tabular}	
}
\vspace{-2mm}
\caption{ \small Distillation experiment with our proposed \emph{Slow} approach as teacher and the \emph{Fast} NCE BoW as student.}\label{tab:distillation_ablation}
\end{table}

\subsection{Thinking Fast and Slow}
\label{sec:rerank}

This section focuses on getting both a fast and accurate model for retrieval.
First, we evaluate the benefit of the distillation from the \emph{Slow} to the \emph{Fast} model.
Next, we evaluate the benefit of the re-ranking strategy and validate our combined approach on a large-scale retrieval experiment.

\noindent \textbf{Distillation improves dual encoder models.}
In Table~\ref{tab:distillation_ablation}, we use our approach, denoted as \emph{Slow}, to distill the knowledge to a \emph{Fast} NCE BoW student dual encoder. 
The distillation improves the performance of the \emph{Fast} model with improvements of over $10\%$ on R@1 when training on COCO, significantly reducing the gap between the \emph{Slow} and \emph{Fast} models. 
On the other hand, the improvements when training on CC are moderate, but we believe the gap can be further reduced by training longer on CC as we found the distillation often takes significantly longer to converge. 

\noindent \textbf{Benefits of re-ranking.}
Table~\ref{tab:rerank_distill} provides the results from re-ranking.
We see that with $K$ as low as $10$, we are able to recover or outperform the performance of the \emph{Slow} model in terms of R@1 while significantly decreasing the query time.
Combining re-ranking with distillation leads to further improvements: on COCO, we can significantly decrease from $K=50$ to $K=10$ the number of examples to re-rank to outperform the \emph{Slow} model thanks to the distillation.
In particular,  we see a 100$\times$ reduction in retrieval time on COCO from our \emph{Slow} to our \emph{Fast \& Slow} (K=10) model.
Note that for the rest of the experimental section, the \emph{Slow} model runs with an increased image resolution of 384 $\times$ 384 for better results,  albeit with slower inference.

\begin{figure}[t]
\centering
\begin{subfigure}{\columnwidth}
  \centering
  \includegraphics[width=\linewidth]{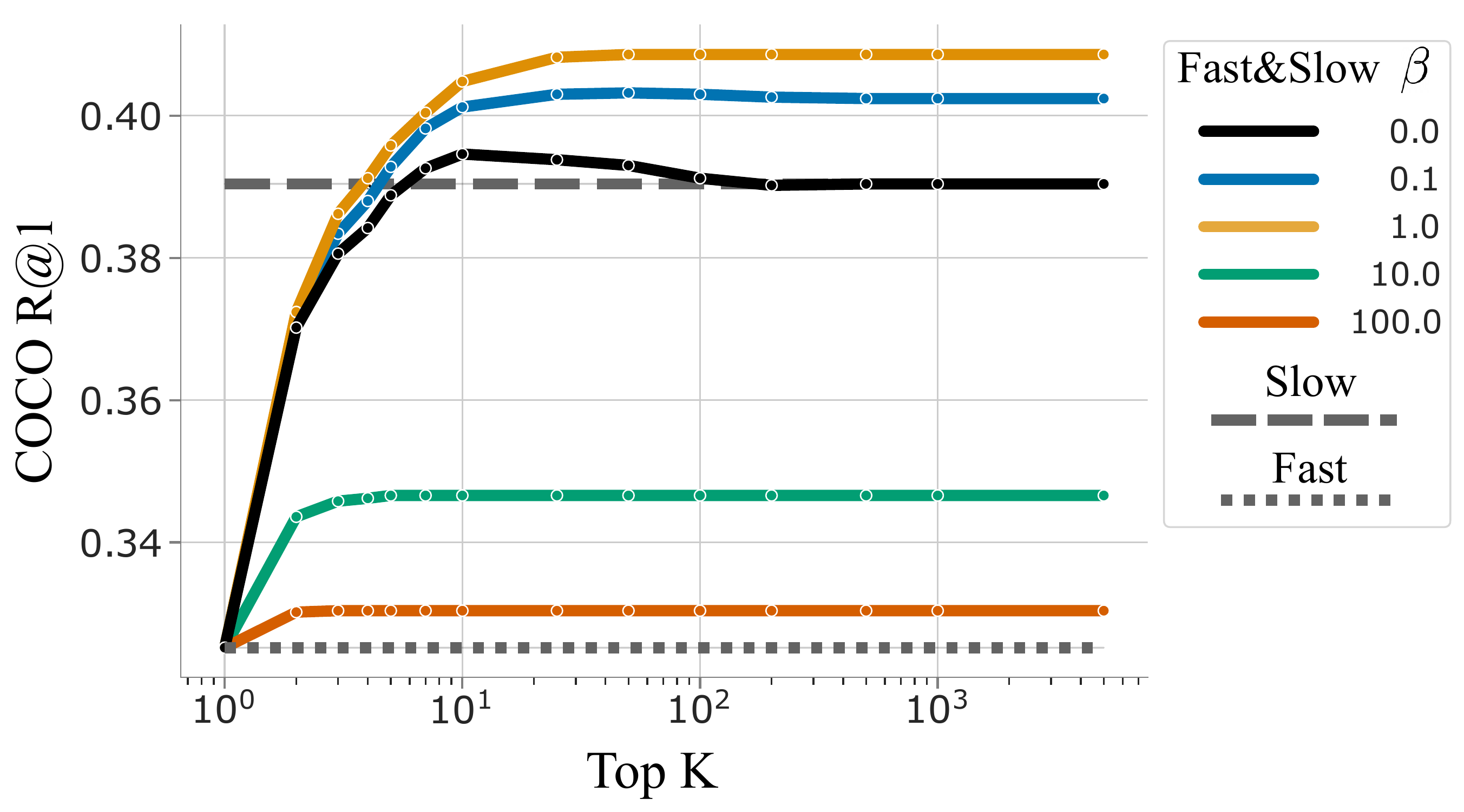}
\end{subfigure} \\ %
\vspace{-3mm}
\caption{\small Retrieval result when varying the top-K retrieved examples from the distilled \emph{Fast} model with varying $\beta$ (See Eq.~\eqref{eq:rerank}).}
\label{fig:rerank_curves}
\end{figure}

Figure~\ref{fig:rerank_curves} provides a more detailed visualization of the effect of re-ranking with respect to the number of top K examples returned from the \emph{Fast} distilled model.
Notably, we see on COCO that \emph{re-ranking as few as five images out of five thousand} from the distilled \emph{Fast} model is enough to reach the \emph{Slow} model R@1 performance.
More quantitative and qualitative results are given in Appendix~\ref{app:additional_results}.

\begin{table}
	\resizebox{\linewidth}{!}{	
		\begin{tabular}{@{}lccccccc|cc@{}}
			Model & Top K & Dist. & Train  & F-R@1 & F-R@5  & C-R@1 & C-R@5 & F-Qt & C-Qt   \\
			\midrule
			\emph{Slow} & \xmark  & \xmark  &     \multirow{5}{*}{COCO}   &   44.8 & 70.4  &  39.0 & 67.7 & 4 s & 19 s\\
			\multirow{4}{*}{\emph{Fast \& Slow}} & 10 & \xmark &         &   44.0    &  63.0  &  38.6     & 61.5  &  \textbf{0.12 s} & \textbf{0.12 s}\\
			& 10 & \cmark &         &   47.2    &  70.1  &  40.5      & 67.8  &  \textbf{0.12 s} & \textbf{0.12 s} \\
			& 50 & \xmark &        &   46.7    &  65.6  &  40.2      & 68.2 &  0.60 s & 0.60 s \\
			& 50 & \cmark &        &     \textbf{47.6}   &   \textbf{73.2} &   \textbf{40.9}   &   \textbf{70.0} &  0.60 s & 0.60 s \\
			\hline
			 \emph{Slow}  & \xmark  & \xmark  &     \multirow{5}{*}{CC}   &  46.9  &  71.5 & 21.0  & 43.3 & 4 s & 19 s\\
			\multirow{4}{*}{\emph{Fast \& Slow}}  & 10 & \xmark &         & 47.7     &  66.6 &  22.6    & 41.1 &  \textbf{0.12 s} & \textbf{0.12 s} \\
			& 10 & \cmark &   & 48.4  & 67.4   &   22.7     & 43.4 &  \textbf{0.12 s} & \textbf{0.12 s} \\
			& 50 & \xmark &   & 50.2    &   73.4    &  \textbf{23.8}  &  \textbf{46.9}     &  0.60 s & 0.60 s \\
			& 50 & \cmark &   &  \textbf{50.5}    &   \textbf{73.6}    &  \textbf{23.8}  &   \textbf{46.9}       &  0.60 s & 0.60 s \\
		\end{tabular}	
	}
	\vspace{-2mm}
	\caption{ \small{Combination of re-ranking and distillation. Dist.: distillation. F-Qt (resp.\ C-Qt) is the query time in seconds on Flickr with 1k images (resp.\ COCO with 5k images) using 1x V100 GPU.}}\label{tab:rerank_distill}
\end{table}

\noindent \textbf{Discussion of scalability.}
We would like to emphasize that the combination of the distillation and re-ranking would be even more appealing in the large-scale retrieval regime as our method allows application of fast approximate nearest neighbour search~\cite{datar2004locality,muja2009fast,sivic03videogoogle,jegou2010product} and hence can potentially scale to billion-scale image retrieval. As a result, our method scales sub-linearly with the number of test images and the time complexity mostly depends on the top $K$, which is the number of calls to the \emph{Slow} model.

\subsection{Comparison to the state of the art}
\label{sec:sota}

We compare to the state of the art on Flickr30K in Table~\ref{tab:sota_image} for the zero-shot and fine-tuning setting.

The \emph{Fast} model is distilled from the \emph{Slow} model on the pretraining dataset (COCO or CC).
The \emph{Fast} model and the \emph{Slow} $384\times 384$ models are then fine tuned on the Flickr30K training set. 
When pretraining on CC, we significantly outperform the VilBERT~\cite{lu2019vilbert} approach despite not using extra information contained in object detectors.
On COCO, we outperform PixelBERT~\cite{huang2020pixel} with the same ResNet-50 backbone while neither training on Visual Genome (VG) annotations nor using high image resolution.
Finally,  we are still below the performance reported in UNITER~\cite{chen2019uniter} and OSCAR~\cite{li2020oscar}.  
We believe this remaining gap can be attributed to \textit{(i)} not using the same amount of pretraining data (UNITER was trained on the combination of four datasets: COCO, CC but also Visual Genome (VG) and SBU and OSCAR is trained on Flickr,  CC,  SBU and GQA~\cite{hudson2019gqa}), \textit{(ii)} not using the same high input image resolution,
\textit{(iii)} not relying on pre-trained object detectors,
and \textit{(iv)} having a smaller model (3 layers transformer with hidden dimension 512 vs.\  24 layers with dimension 1024 for UNITER).
However our proposed approach enables fast retrieval at scale which is not possible out of the box with any of the previously mentioned methods.
More importantly, our scaling approach (distillation and re-ranking) can also be applied to other multimodal transformers including UNITER and OSCAR.

\noindent \textbf{Extension to video.} Our approach can also be applied to video.
To do so, we extend the architecture introduced in Section~\ref{sec:im_text_model} to a TSM ResNet50 model~\cite{lin2019tsm} with the following modifications.
The input of the network is now a sequence of $32$ frames at resolution $224\times 224$. 
Due to memory constraints, we only upsample the last feature map to a $14\times 14$ grid and allow the decoder to attend to the resulting spatio-temporal volume representing the video of shape $32\times 14 \times 14$ (details in Appendix~\ref{app:video_arch}).
We use a pretrained TSM ResNet-50 network~\cite{alayrac2020self} on HowTo100M~\cite{miech2019end2end} and AudioSet~\cite{gemmeke2017audio} datasets.
Results are given in Table~\ref{tab:sota_video}.
We observe that: (i) the upsampling architecture is also beneficial for video, and (ii) our Fast and Slow model sets a new state of the art on this benchmark.

\begin{table}
	\resizebox{\linewidth}{!}{	
		\begin{tabular}{@{}lccp{14mm}cccc@{}}
			Method & Object Det. & Size & Train & Zero-shot & F-R@1 & F-R@5  & F-R@10  \\
			\midrule
			VILBERT~\cite{lu2019vilbert} & \cmark & Full & \multirow{4}{*}{CC} & \multirow{2}{*}{\cmark} & 31.9 & 61.1 & 72.8 \\
			\emph{Fast} and \emph{Slow} (K=100) & \xmark  & 384 & &  & \textbf{48.7} & \textbf{74.2} & \textbf{82.4} \\
			 VILBERT~\cite{lu2019vilbert} & \cmark  & Full &  & \multirow{2}{*}{\xmark} & 58.2  & 84.9 & 91.5 \\
			\emph{Fast} and \emph{Slow} (K=100) & \xmark  & 384  & &  & \textbf{68.2}  & \textbf{89.7} & \textbf{93.9} \\
			\hline
			PixelBERT (R50)~\cite{huang2020pixel} &  \multirow{2}{*}{\xmark}  & 800   & \small{COCO +VG} & \multirow{2}{*}{\xmark}  & 59.8  & 85.5 & \textbf{91.6} \\
			\emph{Fast} and \emph{Slow} (R50, K=100) & & 384 & \small{COCO} &  \xmark & \textbf{62.9} & \textbf{85.8} &  91.3 \\
			\hline
		    Unicoder-VL~\cite{li2019unicodervl}& \cmark  & Full  & \small{CC + SBU} & \xmark & 71.5  & 90.9 & 94.9 \\
		    UNITER~\cite{chen2019uniter}& \cmark  & Full  & \small{COCO +CC +SBU +VG} & \xmark &  75.6  & \textbf{94.1} & \textbf{96.8} \\ 
            OSCAR~\cite{li2020oscar} & \cmark  & Full  & \small{COCO +CC +SBU +GQA} & \xmark & \textbf{75.9}  &  93.3 &  96.6 \\ 
			\emph{Fast} and \emph{Slow} (K=100) & \xmark  & 384 & \small{COCO +CC} &  \xmark  &  72.1 & 91.5 & 95.2 \\
		\end{tabular}	
	}
	\vspace{-2mm}
	\caption{ \small Comparison to state of the art for text-to-image retrieval.  OSCAR results were reproduced from recent work~\cite{geigle2021retrieve}.}\label{tab:sota_image}
\end{table}

\begin{table}
\centering
		\begin{tabular}{@{}lccc@{}}
			Method  & R@1 & R@5  & R@10  \\
			\midrule
			Dual~\cite{dong19dual} & 31.1 & 67.4 & 78.9 \\
			HGR~\cite{chen2020fine} & 35.1 & 73.5 & 83.5 \\
			Support-set~\cite{patrick2020support} & 45.9 & 82.4 & 90.4 \\
			\emph{Fast} NCE BoW   & 42.3 & 79.1 & 88.0 \\
			\emph{Fast} and \emph{Slow (7 $\times$ 7)} (K=10)  & 47.5 & 81.4 & 88.0 \\
			\emph{Fast} and \emph{Slow (14 $\times$ 14)} (K=10)  & \textbf{50.5} & 83.4 & 88.0 \\
			\emph{Fast} and \emph{Slow (14 $\times$ 14)} (K=50)  & \textbf{50.5} & \textbf{84.6} & \textbf{91.7} \\
		\end{tabular}	 
		\vspace{-3mm}
	\caption{ \small Comparison to state of the art retrieval on VATEX.}\label{tab:sota_video}
\end{table}

\section{Conclusion}
We have shown how to scale-up powerful vision-text transformer-based models for retrieval.
In particular, we have introduced an accurate but \emph{Slow} text-vision transformer-based architecture with fine-grained cross-attention for retrieval.
To make it scalable for text-to-visual search,  we have augmented this \emph{Slow} model with a \emph{Fast} dual encoder model through a combination of distillation and re-ranking.
As a result, the combined \emph{Fast \& Slow} approach achieves better results than the \emph{Slow} model while significantly reducing the inference time by several orders of magnitude on large datasets.
We emphasize that our approach is model agnostic and can be applied to any vision-text Transformer \emph{Slow} model and dual-encoder \emph{Fast} retrieval model. 

\vspace{-3mm}
\paragraph{Acknowledgements.}
We would like to thank Lisa Anne Hendricks for  feedback.
The project was partially funded by the French ANR as part of the ``Investissements
d'avenir" program,  reference ANR-19-P3IA-0001 (PRAIRIE 3IA Institute),  and the European Regional 
Development Fund under the project IMPACT (reg.\  no.\
CZ.02.1.01/0.0/0.0/15\_003/0000468).

{\small
\bibliographystyle{ieee_fullname}
\bibliography{master-biblio}
}

\setcounter{section}{0}
\renewcommand{\thesection}{\Alph{section}}

\section{Appendix}
\label{sec:appendix}

We provide here more details about the main paper.
Section~\ref{app:additional_results} gives additional ablation results for the distillation method and the Flickr re-ranking curve (similarly to the COCO re-ranking curve in the main paper).
In Section~\ref{app:optim}, we provide additional details about our training hyperparameters.
Section~\ref{app:arch} describes in more details the proposed architecture for upsampling as well as the video architecture extension.
Finally, in Section~\ref{app:qual_res} we provide qualitative results of our approach.

\section{Additional quantitative results}
\label{app:additional_results}

\noindent \textbf{What matters for good distillation.}
In Table~\ref{tab:ablation_distill_text}, we explore various text models for the \emph{Fast} dual-encoder student when performing distillation.
Interestingly, the BoW model still seems to be the best fit for distillation, hinting that complex language models are not necessarily the most important for the task we consider.
This is in line with our findings in the main paper about the use of more complex language models for \textbf{Noise-Contrastive Estimation (NCE)} (Equation 4 from the main paper) that did not lead to improvements.
Table~\ref{tab:distillation_ablation_temp_combination} shows that care should be given to the choice of temperature used when performing the distillation and that combining the distillation loss with the original loss is crucial to ensuring improvements.
Note that we follow~\cite{hinton2015distilling} and adapt the combination factor $\alpha$ with respect to the temperature parameter $\tau$. 
Based on that study, we use $\tau=10$ and $\frac{\alpha}{\tau^{2}}=0.001$ for all other distillation experiments of the paper.
The models were trained on COCO for 50k steps.

\begin{table}
	\resizebox{\linewidth}{!}{	
		\begin{tabular}{@{}lccccc@{}}
			Text model  & Depth & F-R@1 & F-R@5  & C-R@1 & C-R@5  \\
			\midrule
			Bag-of-words   & 1 &  \textbf{35.6} & \textbf{60.8}  & \textbf{31.2}  & \textbf{61.1}  \\
			\hline
			\multirow{4}{*}{Transformer} & 1 &  27.4 & 51.7 & 20.1  & 45.0 \\
			& 3  &  26.3 & 49.7  &  19.4  & 43.6  \\
			& 6   &  27.1 & 49.9  &  20.0 & 44.0 \\
			& 12   &  28.9 & 50.0 & 19.6 &  43.5 \\
			\hline
			\multicolumn{2}{c}{\textcolor{mygray}{\emph{Slow} upper bound}}  &  \textcolor{mygray}{42.2} & \textcolor{mygray}{66.8} & \textcolor{mygray}{38.5} & \textcolor{mygray}{65.2} \\
	\end{tabular}}
	\caption{ \small Distillation: Which text model to use for the dual encoder approach on COCO.}\label{tab:ablation_distill_text}
\end{table}

\begin{table}
	\resizebox{\linewidth}{!}{	
		\begin{tabular}{@{}cccccc@{}}
			$\tau$  & $\frac{\alpha}{\tau^{2}}$   & F-R@1 & F-R@5  & C-R@1 & C-R@5  \\
			\midrule
			\multirow{4}{*}{1.0}  & 0.0 &  26.8 & 54.0 & 24.1 & 52.7  \\
			& 0.1  &  28.2 & 54.0 & 24.9 & 55.8   \\
			& 1.0  &  27.2 & 51.1 & 25.2 & 52.8   \\
			& 10.0  &  19.1 & 39.6 & 19.0 & 42.6  \\
			\hline
			\multirow{4}{*}{10.0} &  0.0 & 34.0 & 61.6 & 31.1  & 61.0 \\
			&  0.1 & 35.7 & 61.0 & 30.9 & 61.2  \\
			& 1.0   &  34.5 & 61.0 & 31.7 & 61.0   \\
			& 10.0   &  27.2 & 53.4 & 26.2 & 54.6   \\
		\end{tabular}	
	}
	\caption{ \small Distillation temperature and loss weighting ablation study. The models were trained on COCO for 50k steps.} \label{tab:distillation_ablation_temp_combination}
\end{table}

\begin{figure}
\begin{center}
	\centering
	\includegraphics[width=\linewidth]{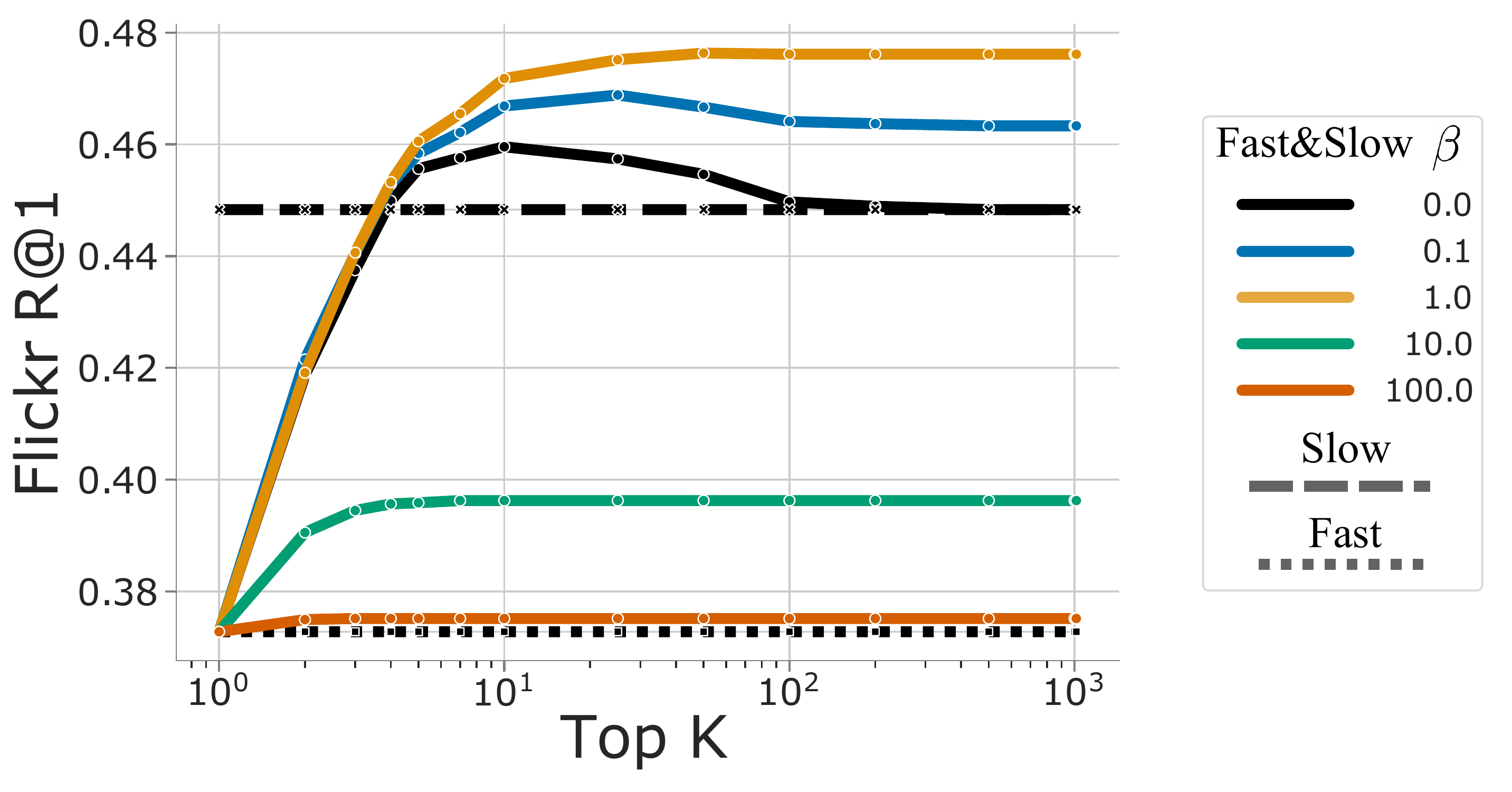}
	\captionof{figure}{Flickr retrieval result when varying the top-K retrieved examples from the distilled \emph{Fast} model with varying $\beta$.}
	\label{fig:rerank_flickr}
\end{center}
\end{figure}

\noindent \textbf{Flickr re-ranking results when varying K.}
Figure~\ref{fig:rerank_flickr} provides a more detailed visualization of the effect of re-ranking with respect to the number of top K examples returned from the \emph{Fast} distilled model on the Flickr validation set. The models were trained on COCO.
Here again, we see that \emph{re-ranking as few as four images out of thousand} from the distilled \emph{Fast} model is enough to reach the \emph{Slow} model R@1 performance.

\section{Experiment details}
\label{app:optim}

\paragraph{Training time data augmentation.}
We randomly crop images using the tensorflow function {\small\texttt{tf.image.sample\_distorted\_bounding\_box\footnote{\url{https://www.tensorflow.org/api_docs/python/tf/image/sample_distorted_bounding_box}}}} with the following parameters: 
\begin{itemize}
\item \texttt{min\_object\_covered=0.2},
\item \texttt{aspect\_ratio\_range=(3 / 4, 4 / 3)},
\item \texttt{area\_range=(0.2, 1.0)}.
\end{itemize} 

The crops are then resized to $224 \times 224$ and randomly flipped from left to right with a probability of $0.5$.
Note that while flipping an image from left to right is a common data augmentation technique in classification, it can be problematic in captioning annotation that mention parts of images specifically to the left or right.
However, we counted that on Conceptual Caption around $1.3\%$ of the captions either mention the word \texttt{left} or \texttt{right} while this percentage is down on COCO to $0.6\%$ which is sufficiently low to not cause a problem.
For our video experiment we use the same spatial augmentation but additionaly subsample temporal clips of $32$ frames at $10$ frame per second ($3.2$ seconds) from the original $10$ seconds clips of VATEX.

\paragraph{Test time data augmentation.}
For all image experiments, we simply take the central crop of the image to perform the retrieval.
For videos, we sample $4$ temporally uniformly clips over the video.
Each clip has $32$ frames that are centrally cropped.
We average the visual-text score $h(x,y)$ over the $4$ clips to obtain the final score.

\begin{figure}[t]
\begin{center}
	\centering
	\includegraphics[width=\linewidth]{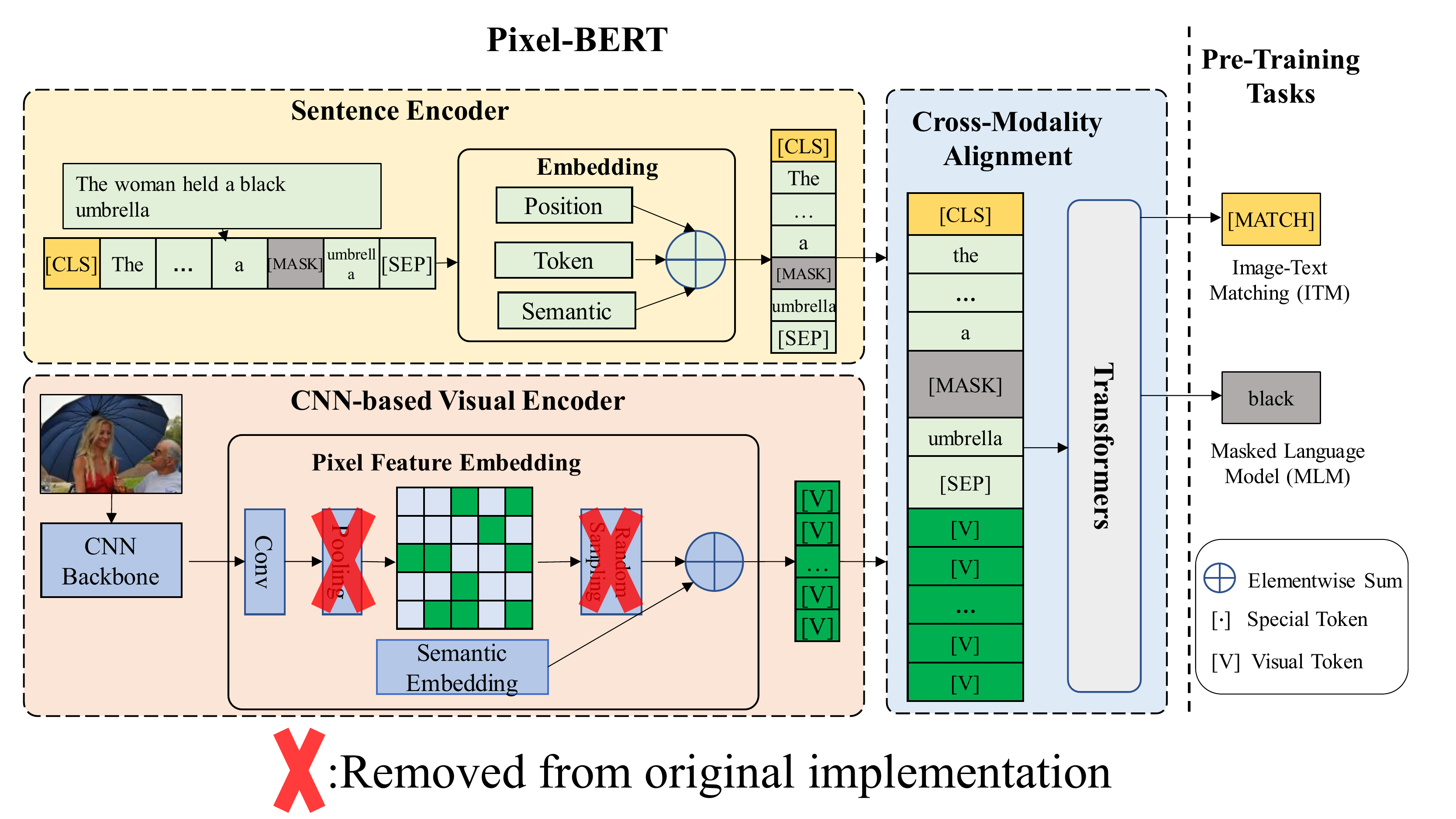}
	\captionof{figure}{Illustration of the PixelBERT~\cite{huang2020pixel}, architecture with the pooling and sampling modules we have removed from the original implementation.}
	\label{fig:pixelbert}
\end{center}
\end{figure}

\paragraph{Training hyper-parameters.}
In the following, we provide the optimization details for each of the trained models on COCO and CC.
We recall that each model is trained using the ADAM optimizer with a cosine learning rate decay and 5k steps of warm up.
\begin{itemize}
\item  \textbf{NCE BoW/BERT}: The models are trained both on COCO and CC using a total batch size of 1024 and a base learning rate of $0.001$. 
On COCO, the model is trained for 20k steps while it is trained for 140k steps on CC. A gradient clip of 30.0 is applied.
When performing distillation, we instead train longer for COCO until 100k steps.
The weight decay is set to 0.0001.
\item \textbf{VirTex and Slow}: The models are trained both on COCO and CC using a base learning rate of $0.0004$. 
On COCO, the model is trained for 250k steps with a batch size of 512 while it is trained for 500k steps  with a batch size of 1024 on CC. A gradient clip of 100.0 is applied.
The weight decay is set to 0.0001.
\item \textbf{PixelBERT}: The models are trained both on COCO and CC using a total batch size of 1024 and a base learning rate of $0.0001$. 
On both CC and COCO, the model is trained for 800k steps. A gradient clip of 30.0 is applied.
The weight decay is set to 0.0001.
Note that because we worked with smaller resolution images compared to the original PixelBERT work, we removed the downsampling block (which follows the ResNet backbone and is composed of a max pooling and random sampling of pixels from the feature map). Instead we feed all the $7 \times 7$ visual features to the transformer. 
Figure~\ref{fig:pixelbert} provides an illustration of the architecture, taken from the original work~\cite{huang2020pixel}, with the modules we have removed to deal with smaller image resolution.
\end{itemize}

The models are trained using the JAX deep learning framework.

\section{Architecture details}
\label{app:arch}

\subsection{Upsampling strategy}

We follow the same upsampling mechanism used in the BiFPN architecture~\cite{tan2020efficientdet}.
In particular we perform the Fast normalized fusion approach:
\begin{align*} 
P^{out} &= \text{SepConv}\left(\frac{w_1\cdot P^{in}  +     w_2 \cdot \text{Resize}(P^{prev})}{w_1 + w_2 + \epsilon}\right) \\
\end{align*}
where $w_i \ge 0$ is ensured by applying a ReLU after each $w_i$, $\epsilon = 0.0001$ is a  small value to avoid numerical instability, \mbox{Resize} is a 2 $\times$ upsampling using a nearest neighbour interpolation, $P^{in}$ is the feature map input of the upsampling block and $P^{prev}$ is the feature map from the previous ResNet feature maps with its dimensionality reduced to 512 through a $1 \times 1$ convolution.
\mbox{SepConv} is a separable depth wise convolution layer followed by a batch normalization and ReLU.
A more detailed illustration of the upsampling architecture is illustrated in Figure~\ref{fig:architecture}.

\subsection{Video architecture with upsampling}
\label{app:video_arch}

\begin{center}
	\centering
	\includegraphics[width=\linewidth]{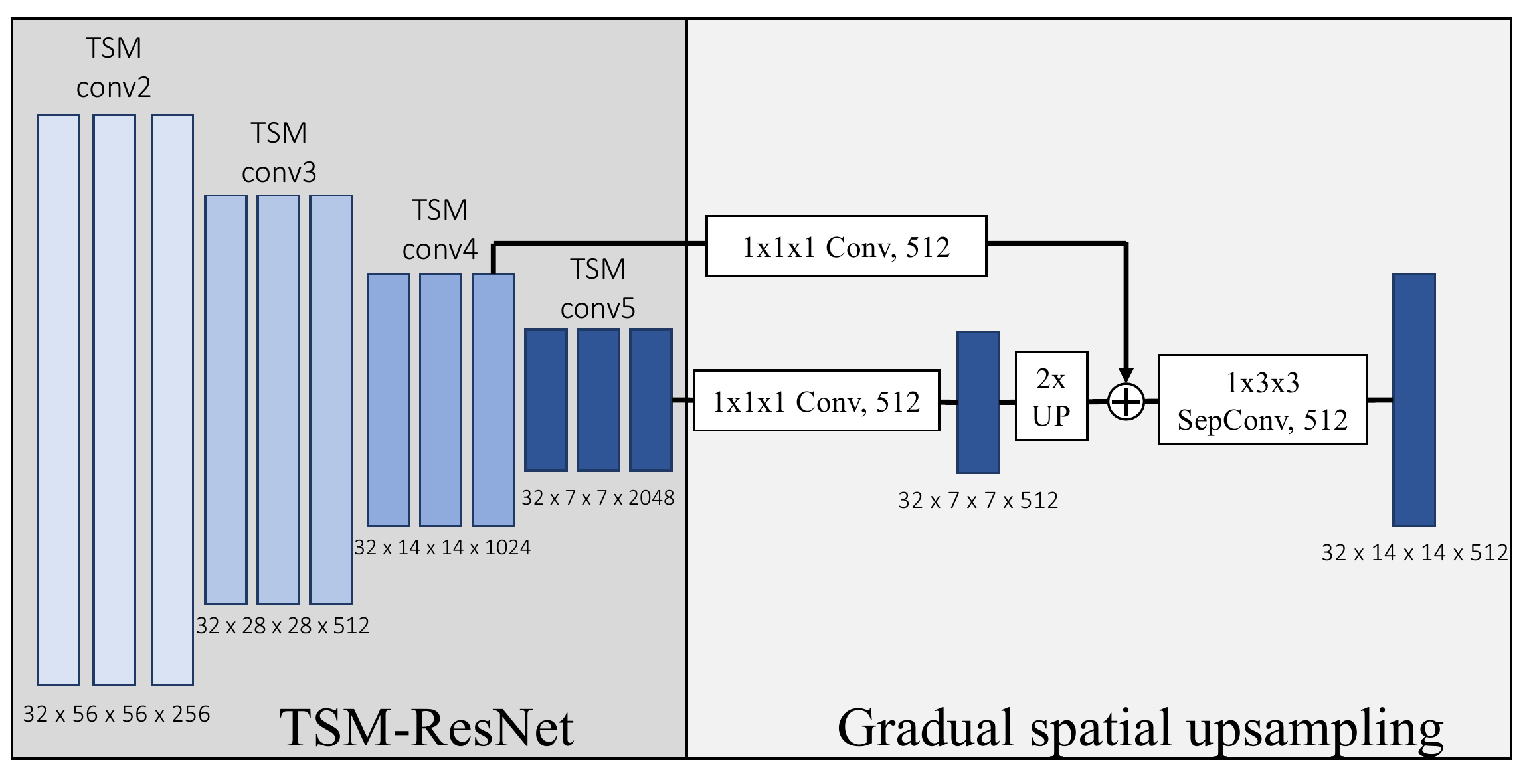}
	\captionof{figure}{\small \textbf{Upsampling architecture for videos:}  adaptation of the upsampling architecture for the TSM network.}
	\label{fig:architecture_tsm}
\end{center}

In order to apply our architecture to video for our experiment on VATEX, we adapt the image only architecture to a video one that can handle spatio-temporal attention.
For that, we start from the TSM ResNet-50 architecture~\cite{lin2019tsm}.
This architecture consists of a standard ResNet-50 model where Temporal Shift Modules (TSM) are inserted within each residual block.
The Temporal Shift Module enables temporal modeling by moving features along the time dimension (forward and backward in time).
One particularity of the model is that there is no temporal pooling hence the temporal resolution stays the same everywhere in the network.
For that reason, we use the same spatial upsampling strategy that we develop for the image network as illustrated in Figure~\ref{fig:architecture_tsm} and simply adapts it to deal with the additional temporal dimension.
Due to memory constraints, we only spatially upsample the feature map to a $14\times14$.
We use clips of $32$ frames sampled at $10$ frame per second for our VATEX experiment ($3.2$ seconds of video).
As a result, the Transformer can attend to a $32\times14\times14$ spatio-temporal feature map.

\section{Qualitative results}
\label{app:qual_res}

\begin{figure}[t]
\begin{center}
	\centering
	\includegraphics[width=\linewidth]{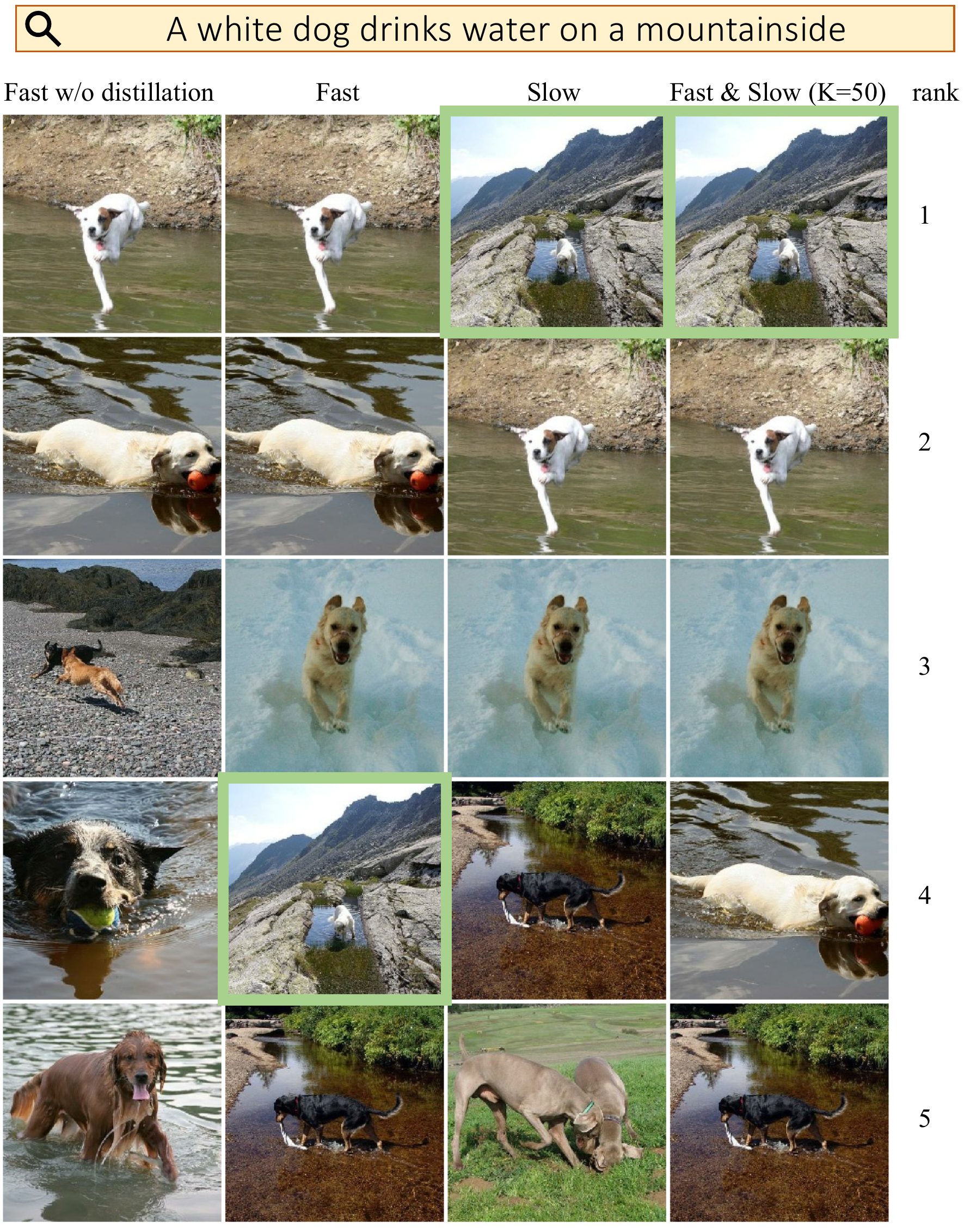}
	\captionof{figure}{\small{Retrieval qualitative examples on Flickr. First column: \emph{Fast} w/o distillation, Second column: \emph{Fast}
, Third column: \emph{Slow} and the last column shows top K=50 re-ranked examples from the \emph{Fast} using the \emph{Slow} model. The image with the green bounding box is the annotated groundtruth for the query.
Note that the groundtruth image is not retrieved in the top-5 with the \emph{Fast} model without distillation and is retrieved at the 4th place with distillation.
The Slow model correctly retrieves it at the first place similarly to the re-ranking model.
Moreover, we can see that the \emph{Slow \& Fast} approach is the only one that either retrieves a white dog or a dog on water.}}
	\label{fig:rerank}
\end{center}
\end{figure}
\subsection{Retrieval results}
We also provide retrieval results on Flickr using four approaches: \emph{Fast} w/o distillation, \emph{Fast}, \emph{Slow} and \emph{Fast and Slow (K=50)}.
Figure~\ref{fig:rerank} illustrates one retrieval example, where we show the top-5 retrieved examples for each model (first row: \emph{Fast} w/o distillation, second row: \emph{Fast}
, third row: \emph{Slow}). The last row shows the top K=50 re-ranked examples from the \emph{Fast} using the \emph{Slow} model.
The models are trained on COCO and evaluated on Flickr in a zero-shot manner.
Note that we have biased the results towards examples that failed for the \emph{Fast w/o distillation} model but are successfully retrieved with re-ranking.

\subsection{Attention maps}
\label{sec:att_map}

\begin{figure*}[t]
	\begin{center}
		\centering
		\includegraphics[width=\linewidth]{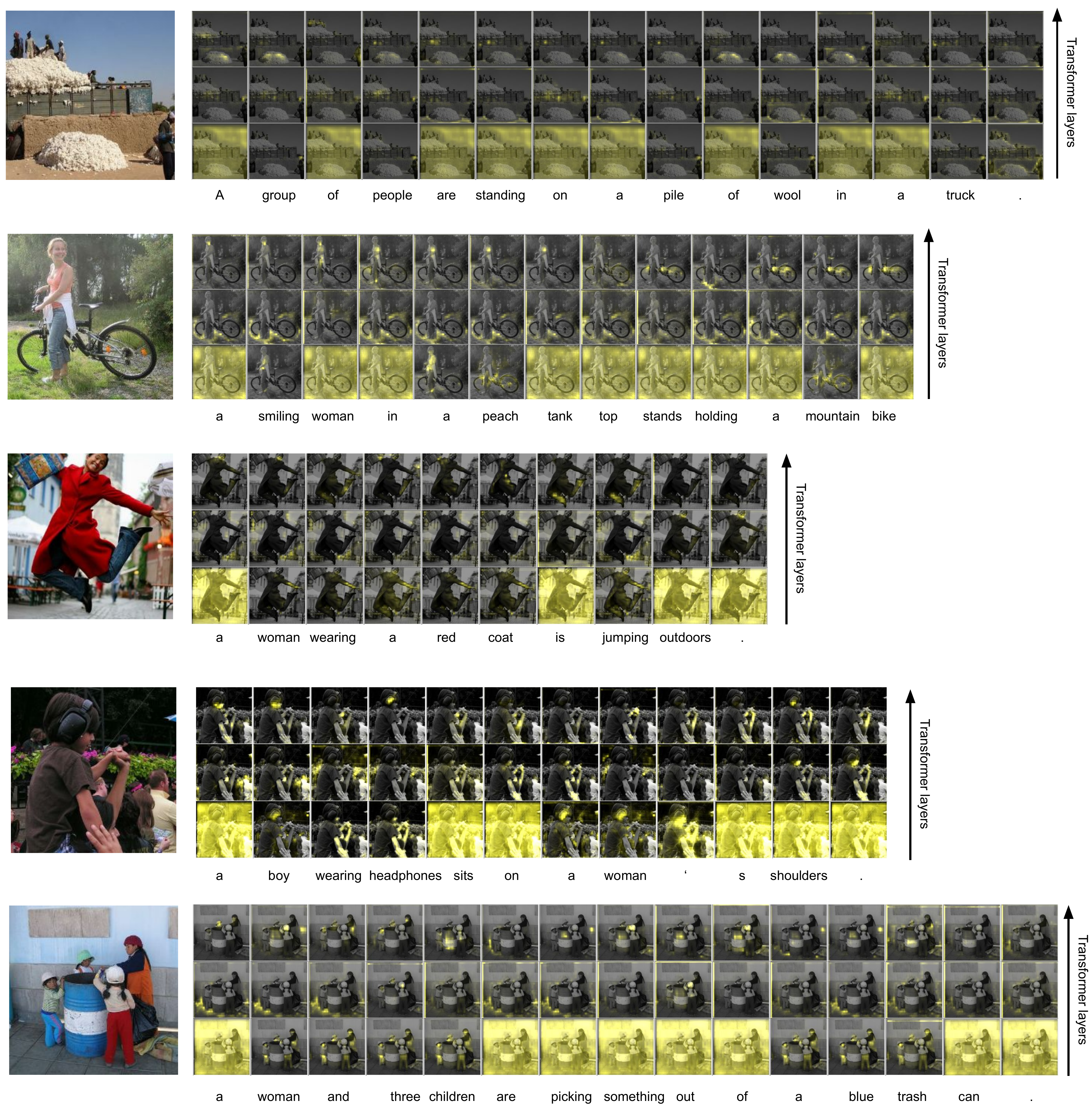}
		\captionof{figure}{\small{Attention maps visualization for the \emph{Slow} model with the $56\times56$ upsampled feature map. For each image, the attention maps are given so that the bottom row corresponds to the  transformer layer closest to the input text. See main text in Appendix~\ref{sec:att_map} for details. Best seen in color on a screen.}}
		\label{fig:56x56}
	\end{center}
\end{figure*}

\begin{figure*}[t]
	\begin{center}
		\centering
		\includegraphics[width=\linewidth]{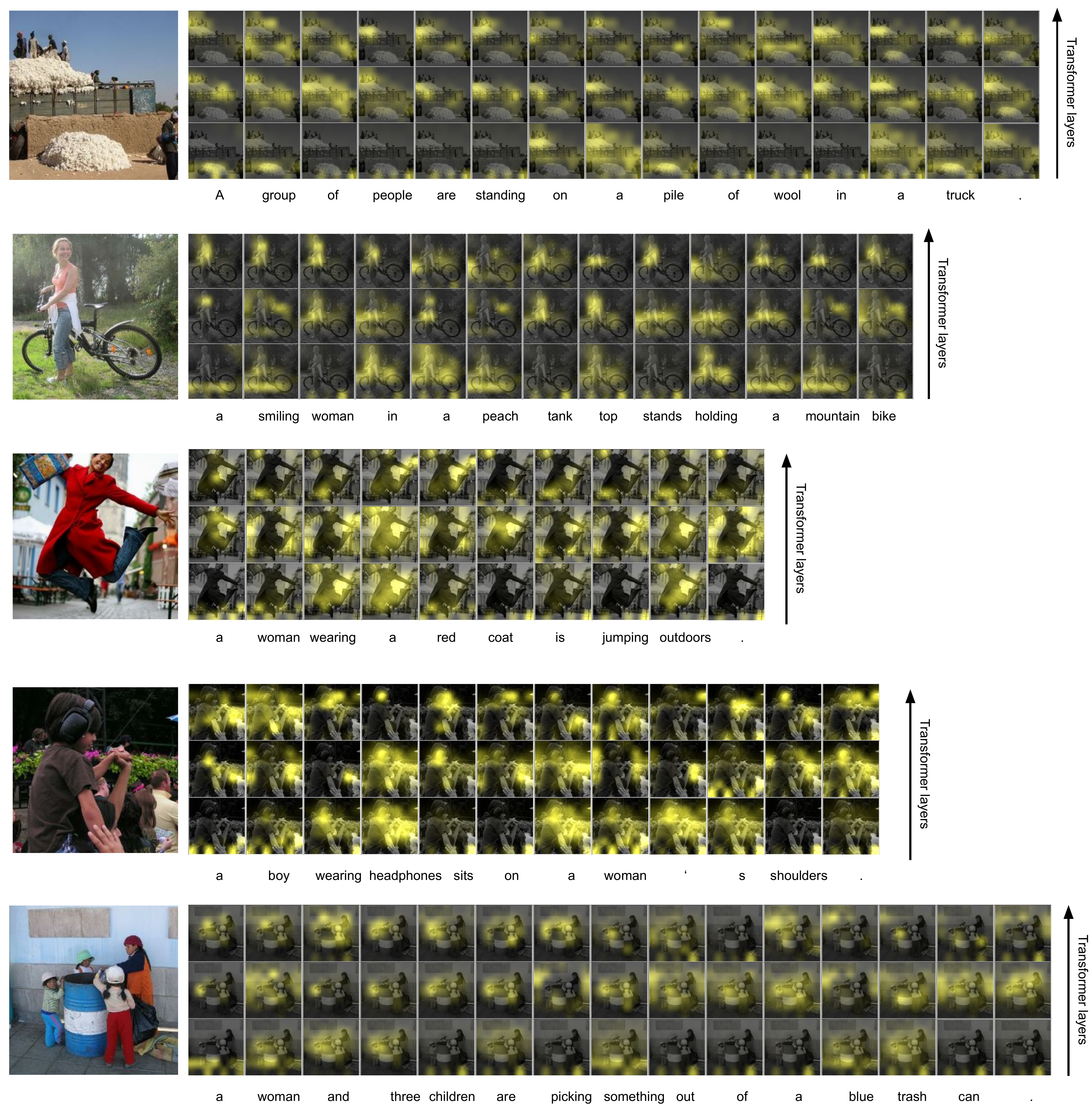}
		\captionof{figure}{\small{Attention maps visualization for VirTex ($7\times7$ attention map). For each image, the attention maps are given so that the bottom row corresponds to the  transformer layer closest to the input text. See main text in Appendix~\ref{sec:att_map} for details. Best seen in color on a screen.}}
		\label{fig:7x7}
	\end{center}
\end{figure*}

In this section, we provide a qualitative analysis of the attention maps between the text and the input image in the Transformer model.
We start by describing how we obtain these attention maps and how we display them.
Next, we provide our main observations and findings from analyzing these feature maps.
To conduct this qualitative study, we compare two models trained on COCO: \textbf{(i)} our \emph{Slow} model (see Figure~\ref{fig:56x56}) which can attend to the $56\times56$ upsampled feature map and \textbf{(ii)} our reimplementation of the VirTeX model (see Figure~\ref{fig:7x7}) with decoder heads attending to a $7\times7$ feature map.

\paragraph{Extracting and visualizing attention maps.}
Recall that our textual decoder is a 3 layer Transformer.
Each of these layers can perform cross-attention between an input word token and the whole image feature map of size $H\times H$ in order to output prediction scores for the next word.
In detail, at each layer and for each input text token, a \emph{query} text vector is produced and compared to the precomputed $H\times H$ visual \emph{keys} via dot product.
The resulting unnormalized $H\times H$ scores are then normalized with a softmax.
These normalized weights are then used to aggregate $H\times H$ visual vectors, or \emph{values}, that are used to update the current token representation in order to predict the scores of the next word.
This process is done in parallel for $8$ attention heads before concatenating the outputs of all heads in a single vector.

This gives an opportunity to visualize the attention maps to see what are the regions of the images that are attended to in order to output a given word.
We provide such visualization in Figure~\ref{fig:56x56} and Figure~\ref{fig:7x7}.
Each figure has multiple examples shown in different rows.
On the left, we show the input image in color.
On the bottom right, we show the input tokens shifted backward by one word so that we direclty visualize what the attention maps look like for the prediction of the current word.
On the right, we show the attention maps for the different layers of the transformer in yellow overlaid over the gray image.
The attention map rows are ordered so that the bottom one corresponds to the layer closest to the input text.
The attention maps are resized to the original input image resolution ($224\times224$) via bicubic sampling.
We only show a single attention head per token and per layer by selecting the one that has the highest average score over the $H\times H$ grid prior to applying the softmax.
For that study, we use as inputs the ground truth captions and images from the Flickr validation set in order to emulate what happens when the Transformer model is used for retrieval.

\paragraph{Analysis.}
Looking at Figure~\ref{fig:56x56} and Figure~\ref{fig:7x7} we make the following observations.

First of all, in both cases we see that there is some level of coherency between the attended regions and the tokens being predicted.
For example, in the second row of Figure~\ref{fig:56x56} (the woman with the bicycle), we see that the last layer of the  transformer attends to the mouth of the woman for the word ``smilling'', at the shirt of the woman to predict the color ``peach'' of the top, and finally attends to the region containing the bicycle to predict the last word.

Second, we observe that, as expected, the attention maps obtained from our \emph{Slow} model in Figure~\ref{fig:56x56} is indeed fine grained when compared to the original VirTeX attention maps of Figure~\ref{fig:7x7}. 
This can notably be seen on the last row of the figures, where the \emph{Slow} model can attend more precisely to the faces of the children thanks to the higher $56\times56$ resolution feature map compared to the crude $7\times7$ feature map of the original VirTeX.

Third, we note that for the \emph{Slow} model, the attention becomes more focused for the higher layers that are closer to the output.
This is notably true when being input the first word where the attention systematically covers the full image for the first layer before being refined on a specific region in the image.

Finally, while these visualizations are not always perfectly interpretable, we believe similar studies and inspections are valuable to better understand how these models relate text and vision.

\end{document}